%% file: ARTICLE.tex
\documentclass[letterpaper, 10 pt, conference]{ieeeconf}

\IEEEoverridecommandlockouts
\usepackage[english]{babel}
\usepackage{amsmath}
\usepackage{amssymb}
\usepackage{booktabs}
\usepackage[dvips]{graphicx}
\usepackage{multirow}
\usepackage{amsfonts}
\usepackage{enumerate}
\usepackage{tabularx}
\usepackage{algorithm,algorithmic}
\usepackage{bm}
\usepackage{enumerate}
\usepackage{adjustbox}
\usepackage{xcolor}
\usepackage{siunitx}
\usepackage{booktabs}
\usepackage{mdframed}
\usepackage{adjustbox}
\usepackage{authblk}
\usepackage{color}
\usepackage{xurl}
\usepackage{subcaption}
\usepackage{cite}
\makeatletter
\let\NAT@parse\undefined
\makeatother
\usepackage{hyperref}
\usepackage[toc,sort=use]{glossaries}
\usepackage{array}
\newglossary[slg]{symbols}{sym}{sbl}{List of Symbols}
\newglossary[slg]{hidden}{sym}{sbl}{Hidden Symbols}
\makeglossaries
\input{symbol_list}

\input{notation}

\usepackage{longtable}              
\makeatletter
\let\saved@longtable\longtable
\long\def\foo#1\LT@err#2#3#4!!{\def\longtable{#1#4}}
\expandafter\foo\longtable!!

\long\def\foo#1\@outputpage#2\@outputpage#3!!{%
	\def\LT@output{#1\@opcol#2\@opcol#3}}
\expandafter\foo\LT@output!!
\makeatother

\newglossarystyle{symboltable}{%
	{
		\begin{table} %[!h]
			\caption{List of Symbols}	
			\label{table:list_symbols}	
			\begin{center}	
				\begin{tabular}{|l|m{6.5cm}|}		
				}{
				\hline
			\end{tabular}
		\end{center}
	\end{table}	
}%
\renewcommand{\glossarysection}[2][]{} % no title
\renewcommand
*
{\glossaryheader}{
	\hline
	\textbf{Symbol}
	&
	\textbf{Description}
	\\\hline
}%
\renewcommand
*

{\glsgroupheading}[1]{}%
\renewcommand
*
{\glsgroupskip}{}%
\renewcommand
*
{\glossentry}[2]{%
	\glstarget{##1}{\glossentryname{##1}}% the entry name
	&
	\space \glossentrydesc{##1}% the description
	.
	\\
}%
\renewcommand
*
{\subglossentry}[3]{%
	\glossentry{##2}{##3}}%
}

 \usepackage{cite}

\usepackage{wasysym}  % symbols
\author{
    Arpan Pallar$^{*}$, Guanrui Li$^{*}$, Mrunal Sarvaiya, and Giuseppe Loianno
\thanks{
    *{Equal contributon}.}
\thanks{The authors are with the New York University, Tandon School of Engineering, Brooklyn, NY 11201, USA. {\tt\footnotesize email: \{ap7538, lguanrui, mrunal.s, loiannog\}@nyu.edu}.}
\thanks{This work was supported by the NSF CPS Grant CNS-2121391, the NSF CAREER Award 2145277, Qualcomm Research, Nokia, and NYU Wireless.}
\thanks{The authors acknowledge Manling Li for her help and support on the robot design, manufacture and experiments.}
}

\title{\LARGE \bf Optimal Trajectory Planning for Cooperative Manipulation with Multiple Quadrotors Using Control Barrier Functions}

\begin{document}
\maketitle
\thispagestyle{empty}
\pagestyle{empty}

\begin{abstract}

In this paper, we present a novel trajectory planning algorithm for cooperative manipulation with multiple quadrotors using control barrier functions (CBFs). Our approach addresses the complex dynamics of a system in which a team of quadrotors transports and manipulates a cable-suspended rigid-body payload in environments cluttered with obstacles. The proposed algorithm ensures obstacle avoidance for the entire system, including the quadrotors, cables, and the payload in all six degrees of freedom (DoF). We introduce the use of CBFs to enable safe and smooth maneuvers, effectively navigating through cluttered environments while accommodating the system's nonlinear dynamics. To simplify complex constraints, the system components are modeled as convex polytopes, and the Duality theorem is employed to reduce the computational complexity of the optimization problem. We validate the performance of our planning approach both in simulation and real-world environments using multiple quadrotors. The results demonstrate the effectiveness of the proposed approach in achieving obstacle avoidance and safe trajectory generation for cooperative transportation tasks.

\end{abstract}
\section*{Supplementary material}
\textbf{Video}: \url{https://youtu.be/CMVBYsQFwTU}
\IEEEpeerreviewmaketitle

\input{sections/01-Intro}

\input{sections/02-Related_Works}
\input{sections/03-Dynamics} 
\input{sections/04-Planning}
\input{sections/05-Experiments}

\input{sections/06-Conclusion}

\bibliographystyle{IEEEtran}	% (uses file "plain.bst")
\bibliography{references}
%\biboptions{sort&compress}
%\input{10-old.tex}
\end{document}

%% file: symbol_list.tex
\DeclareMathAlphabet{\mathpzc}{OT1}{pzc}{m}{it}
 
\newcommand{\glo}[3]{
	\newglossaryentry{#1}
	{
	  name=#2,
	  description=#3,
	}
}
\newcommand{\glom}[4]{
	\newglossaryentry{#1}
	{
	  name={\ensuremath{#2}},
	  symbol={\ensuremath{#3}},
	  description=#4,
	  type=symbols,
	}
}
\newcommand{\glomh}[4]{
	\newglossaryentry{#1}
	{
	  name={\ensuremath{#2}},
	  symbol={\ensuremath{#3}},
	  description=#4,
	}
}
\glo{naiive}{naive}{ratio of circumference of circle to its
               diameter}

\glomh{linx}{A}{A}{Input matrix.}
\glomh{linp}{C}{C}{parameter matrix.}

\glom{sgn}{\chi}{\chi}{Indicator function}
\glom{error}{e}{e}{Desired roll}
\glom{roll}{\phi}{\phi}{roll}
\glom{pitch}{\theta}{\theta}{pitch}
\glom{yaw}{\psi}{\psi}{yaw}
\glom{droll}{ p}{ p}{roll}
\glom{dpitch}{ q}{ q}{pitch}
\glom{dyaw}{ r}{ r}{yaw}
\glom{angles}{\mathbf{\Theta}}{\mathbf{\Theta}}{Robot attitude}
\glom{dangles}{\mathbf{\Omega}}{\mathbf{\Omega}}{Angular velocities}
\glom{id}{k}{k}{general id of robot used in description of control method and estimation}

\glom{thrust}{F}{F}{Thrust}
\glom{u}{\mathbf{u}}{\mathbf{u}}{control input}
\glom{w}{\mathbf{w}}{\mathbf{w}}{control input}
\glom{angular}{\omega}{\omega}{angular speed}

\glom{worldf}{\mathcal{I}}{\mathcal{I}}{World frame}
\glom{loadf}{\mathcal{L}}{\mathcal{L}}{Load frame}
\glom{robotf}{\mathcal{B}}{\mathcal{B}}{Body frame}
\glom{camf}{\mathcal{C}}{\mathcal{C}}{Camera frame}

\glom{tag}{\mathpzc{t}_p}{\mathpzc{t}_p}{Tag}
\glom{robots}{\mathcal{R}}{\mathcal{R}}{Set of a team of robots}
\glom{robotl}{\mathbf{x}}{\mathbf{x}}{Robot location.}
\glom{robot}{\mathbf{x}}{\mathbf{x}}{Robot location.}
\glom{width}{\mathpzc{w}}{\mathpzc{w}}{Robot width.}
\glom{height}{\mathpzc{h}}{\mathpzc{h}}{Robot height.}
\glom{orientations}{\Psi}{\Psi}{Robot orientations.}
\glom{orientation}{\psi}{\psi}{Robot orientation.}
\glom{structure}{\mathit{S}}{\mathit{S}}{Structure}
\glom{structureset}{\mathcal{S_F}}{\mathcal{S_F}}{Structure set}
\glom{component}{\mathit{S}}{\mathit{S}}{Structure}

\glom{targets}{\mathcal{Z}}{\mathcal{Z}}{Target points}
\glom{targetse}{\mathcal{Z}^E}{\mathcal{Z}^E}{Expanded target points}
\glom{target}{z}{z}{Robot location.}

\glom{tree}{\mathcal{T}}{\mathcal{T}}{Assembly tree.}
\glom{tedges}{\mathcal{E}^\mathcal{T}}{\mathcal{E}^\mathcal{T}}{Edges of the Assembly tree.}
\glom{tvertices}{\mathcal{V}^\mathcal{T}}{\mathcal{V}^\mathcal{T}}{Vertices of the Assembly tree.}
\glom{tivertices}{\mathcal{V}^{\mathcal{T}_i}}{\mathcal{V}^{\mathcal{T}_i}}{Vertices of the Assembly tree.}

\glom{graph}{\mathcal{G}}{\mathcal{G}}{Communication graph}
\glom{cgraph}{\hat{\mathcal{G}}}{\hat{\mathcal{G}}}{Canonical robust graph}
\glom{vertices}{\mathcal{V}}{\mathcal{V}}{Set of Vertices}
\glom{neighbors}{\mathcal{N}_i}{\mathcal{N}}{Neighbors for node $i$}
\glom{edges}{\mathcal{E}}{\mathcal{E}}{Set of Edges}
\glom{neighborsi}{\mathcal{J}_i}{\mathcal{J}}{Inclusive Neighbors for node $i$}
\glom{removed}{\mathcal{R}}{\mathcal{R}}{Removed nodes}
\glom{normal}{\mathcal{N}}{\mathcal{N}}{Normal nodes}
\glom{strongset}{\mathcal{F}}{\mathcal{F}}{Strong nodes}
\glom{weakset}{\mathcal{W}}{\mathcal{W}}{Weak nodes}
\glom{candidates}{\mathcal{C}}{\mathcal{C}}{Candidate points}
\glom{boundary}{\mathcal{P}}{\mathcal{P}}{Objective points}

\glom{loc}{v}{v}{Node location in the space}
\glom{val}{x}{x}{Variable of interest of the node}

\glom{received}{\mathcal{P}_i^t}{\mathcal{P}}{Set of received values by node $i$ a the time interval 
}

%% file: notation.tex
\newcommand{\id}{k}

\newcommand{\vecb}{\mathbf{b}}

\newcommand{\vece}{\mathbf{e}}

\newcommand{\vecg}{\mathbf{g}}

\newcommand{\vecq}{\mathbf{q}}

\newcommand{\vecx}{\mathbf{x}}
\newcommand{\vecy}{\mathbf{y}}

\newcommand{\vecmu}{\bm{\mu}}

\newcommand{\vecrho}[1]{\bm{\rho}_{#1}}

\newcommand{\hatvecrho}[1]{\hat{\bm{\rho}}_{#1}}
\newcommand{\vecxi}{\bm{\xi}}

\newcommand{\matA}{\mathbf{A}}
\newcommand{\matB}{\mathbf{B}}

\newcommand{\matF}{\mathbf{F}}

\newcommand{\matI}{\mathbf{I}}

\newcommand{\matM}{\mathbf{M}}

\newcommand{\matQ}{\mathbf{Q}}
\newcommand{\matP}{\mathbf{P}}
\newcommand{\matR}{\bm{\mathit{R}}}
\newcommand{\matU}{\mathbf{U}}
\newcommand{\matV}{\mathbf{V}}
\newcommand{\matW}{\mathbf{W}}
\newcommand{\matX}{\mathbf{X}}

\newcommand{\inertia}{\mathbf{J}}
\newcommand{\inertiaload}{\mathbf{J}_{L}}
\newcommand{\loadmass}{m_{L}}
\newcommand{\half}{\frac{1}{2}}

\newcommand{\angvel}{\mathbf{\Omega}}
\newcommand{\angacc}{\dot{\angvel}}

\newcommand{\robotpos}[1]{\vecx_{#1}}

\newcommand{\robotrot}[1]{\matR_{#1}}

\newcommand{\robotvel}[1]{\dot{\vecx}_{#1}}
\newcommand{\robotacc}[1]{\ddot{\vecx}_{#1}}
\newcommand{\robotangvel}[1]{\angvel_{#1}}

\newcommand{\tension}[1]{\vecmu_{#1}}

\newcommand{\loadpos}{\vecx_{L}}
\newcommand{\loadquat}{\vecq_{L}}
\newcommand{\loadquatdot}{\dot{\vecq}_{L}}

\newcommand{\loadrot}{\matR_{L}}

\newcommand{\loadvel}{\dot{\vecx}_{L}}

\newcommand{\loadacc}{\ddot{\vecx}_{L}}

\newcommand{\loadangvel}{\angvel_{L}}

\newcommand{\loadangacc}{\angacc_{L}}

\newcommand{\cablevec}[1]{\vecxi_{#1}}

\newcommand{\realnum}[1]{\mathbb{R}^{#1}}
\newcommand{\SOthree}{SO(3)}

\newcommand{\twonorm}[1]{\left\lVert#1\right\rVert_2}
\newcommand{\prths}[1]{\left(#1\right)}

\newcommand{\worldf}{\mathcal{I}}
\newcommand{\robotf}[1]{\mathcal{B}_{#1}}
\newcommand{\loadf}{\mathcal{L}}
\newcommand{\axis}[2]{\mathbf{e}_{#1}^{#2}}

\newcommand{\netplforce}{\mathbf{F}}

\newcommand{\netplmoment}{\mathbf{M}}

%% file: sections/01-Intro.tex
\section{Introduction}

Low-cost autonomous Micro Aerial Vehicles (MAVs) equipped with manipulation mechanisms have significant potential for assisting humans with complex and hazardous tasks, such as construction, delivery, and inspection. In construction scenarios, teams of MAVs can collaboratively transport materials from the ground to upper floors, accelerating the construction process. Similarly, in urban settings, where ground traffic can cause delays during rush hour, MAVs can utilize unobstructed airspace to facilitate prompt humanitarian missions or emergency medical deliveries. These tasks necessitate aerial robots that can transport and manipulate objects effectively.

Various mechanisms can be employed to transport and manipulate a payload using a team of aerial robots, including ball joints~\cite{tagliabue2019ijrr,Loiannodesert2018}, robot arms~\cite{jongseok2020icra}, and cables~\cite{guanrui2021iser,klausen2020passiveoutdoortransportation}, as discussed in~\cite{ollero2021TRO}. Cable mechanisms are particularly advantageous due to their lighter weight, lower cost, simpler design requirements, and zero-actuation-energy consumption, making them especially suitable for Size, Weight, and Power (SWaP) aerial platforms~\cite{guanrui2021ral}.

\begin{figure}[t]
\centering
  \includegraphics[width=\columnwidth]{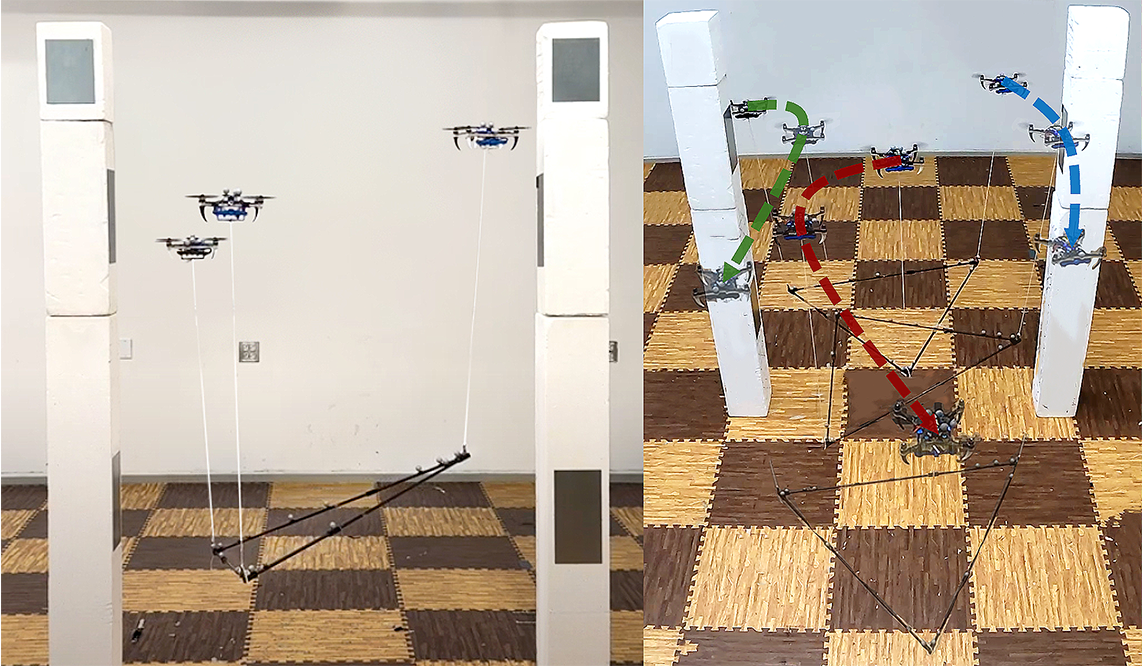}  
  \caption{Multiple quadrotors manipulating a payload to move through narrow opening.\label{fig:null_space_visualization}}
  \vspace{-20pt}
\end{figure}

%TODO decide if this figure looks better than fig 1
However, these systems present significant challenges in developing planning strategies due to indirect load actuation, intricate configuration spaces, and the complex, tightly coupled nonlinear dynamics between the load and the MAVs. While considerable research has been conducted on control strategies, there is a notable gap in the literature regarding planning, particularly for obstacle avoidance of the entire system, including the quadrotor, cables, and rigid-body payload. Therefore, this paper proposes an innovative planning algorithm that generates an obstacle-free trajectory for a team of quadrotors manipulating a cable-suspended rigid-body payload. The proposed algorithm not only ensures obstacle avoidance for the payload in all six degrees of freedom (DoF) but also accounts for the obstacles in the path of the robots and cables, while accommodating the system's inherent complex dynamics.

The contributions of this paper are summarized as follows:
\begin{enumerate}
\item We present a novel planning algorithm that explicitly considers obstacle avoidance for a team of quadrotors transporting and manipulating a rigid-body payload in all 6 DoF. The algorithm ensures that obstacles are avoided not only by the payload but also by the suspended cables and the quadrotors, addressing the full complexity of the system.
\item We introduce the use of Control Barrier Functions (CBFs) to enable smooth and safe obstacle avoidance maneuvers. This approach guarantees safe navigation in cluttered environments.
\item To simplify the complex constraints of the quadrotor-cable-payload system, we model the components as convex polytopes. We further employ the Duality theorem to reduce the complexity of the optimization problem, making it more computationally efficient and feasible for trajectory generation in constrained spaces.
\end{enumerate}

The rest of this paper is organized as follows. In Section \ref{sec:conclusion}, we introduce the system modeling, including the dynamics of the payload, quadrotors, and cables. Section~\ref{sec:trajectory_generation} introduces the proposed trajectory planning algorithm, outlining the cost function and Control Barrier Function (CBF) constraint formulation. Section ~\ref{sec:experimental_results} describes the experimental setup, including both simulation and real-world results, which validate the performance of the algorithm. Finally, Section~\ref{sec:conclusion} concludes the paper, summarizing the contributions and discussing potential future research directions.

%% file: sections/02-Related_Works.tex
\section{Related Works}
Multi-robot cooperative payload transportation is a challenging task, particularly when navigating cluttered environments and avoiding obstacles. Traditional trajectory generation methods for single-robot payload systems, such as \cite{son2020ral}, formulate the problem as an optimization program. However, extending these methods to multi-robot systems is non-trivial due to the increased size of the action space, non-convex form factor of the physical system and inherent redundancy in tracking a desired payload pose \cite{li2023nonlinear}. Early research in this area addressed these challenges by simplifying the problem, assuming static equilibrium conditions at intermediate waypoints, and utilizing optimization solvers to generate conservative trajectories \cite{michael2011cooperative} \cite{fink2011planning}.

The next wave of research focused on integrating obstacle avoidance into trajectory generation by using traditional graph search-based methods. Works like~\cite{lee2015rsj,arab2021ast} employ RRT$^{*}$ to ensure collision-free trajectories, however their experimental results are limited to simple simulation environments. The authors in~\cite{kim2017icra} combine RRT$^*$ with Parametric Dynamic Movement Primitives (PDMPs) but assume constant obstacle widths and only demonstrated real-world experiments with two robots in simple collision avoidance tasks. \cite{lee2018tase} tackles the problem by combining RRT$^*$ with Dynamic Movement Primitives (DMPs) and imposed a leader-follower structure with two robots. 
\cite{pizetta2019isat} proposes a controller based on potential fields to maintain safe inter-robot distances and avoid obstacles. However, they prioritize minimizing payload oscillations, reducing the system's agility, and only showed experiments in simulation. Similarly, \cite{aliyu2022ieeea} develops a controller that employs a potential field for collision avoidance but assumed that the quadrotors maintain a constant formation throughout the trajectory. 
Other works~\cite{li2023nonlinear,li2022safety} mathematically demonstrate how the system's null space can be leveraged to accomplish additional tasks, such as maintaining inter-robot separation and avoiding obstacles while tracking a desired payload pose. While their experiments successfully enforce inter-robot separation, they do not demonstrate obstacle avoidance.

\cite{wang2023ieeesmc} incorporates Control Barrier Functions (CBFs) and Control Lyapunov Functions (CLFs) within a switching controller framework to avoid obstacles. They simplify the collision constraints to reduce computation complexity by only considering the robot closest to an obstacle. Similarly, \cite{hedge2022ieeecsl} uses a CBF for obstacle avoidance but restricted its approach to planar scenarios in simulation.

To the best of the authors' knowledge, no existing trajectory generators explicitly account for collisions involving each component of the multi-robot system, particularly collisions between obstacles, individual robots, and the cable mechanism. Additionally, current methods simplify the problem by assuming a leader-follower structure or a constant quadrotor formation. This work addresses these limitations through a novel optimization formulation that incorporates Control Barrier Functions (CBFs) and demonstrates real-world experiments with three quadrotors in a challenging obstacle avoidance task. 

%% file: sections/03-Dynamics.tex
\begin{table}[t]
\caption {Notation table\label{tab:notation}} 
\centering
%\newcolumntype{s}{}
\begin{tabularx}{0.48\textwidth}{>{\hsize=0.58\hsize}X >{\hsize=1.42\hsize}X}
    \hline\hline
 $\worldf$, $\loadf$, $\robotf{\id}$ & inertial frame, payload frame, $\id^{th}$ robot frame\\
 $\loadmass,m_{\id}\in\realnum{}$ &  mass of payload, $\id^{th}$ robot\\
 $\loadpos,\robotpos{\id}\in\realnum{3}$ &  position of payload, $\id^{th}$ robot in $\worldf$\\
 $\loadvel, \loadacc\in\realnum{3}$ & linear velocity, acceleration of payload in $\worldf$\\
$\robotvel{\id}, \robotacc{\id}\in\realnum{3}$ &linear velocity, acceleration of $\id^{th}$ robot in $\worldf$\\
 $\loadrot\in\SOthree$&orientation of payload with respect to $\worldf$ \\
 $\robotrot{\id}\in\SOthree$&orientation of $\id^{th}$ robot with respect to $\worldf$ \\
 $\loadangvel$, $\loadangacc\in\realnum{3}$ & angular velocity, acceleration of payload in $\loadf$\\
  $\robotangvel{\id}\in\realnum{3}$& angular velocity of $\id^{th}$ robot in $\robotf{\id}$\\
  $f_{\id}\in\realnum{}$, $\matM_{\id}\in\realnum{3}$&total thrust, moment at $\id^{th}$ robot in $\robotf{\id}$.\\
$\inertiaload,\inertia_{\id}\in\realnum{3\times3}$   &  moment of inertia of payload, $\id^{th}$ robot\\%\hline
 $\cablevec{\id}\in S^2$&unit vector from $\id^{th}$ robot to attach point in $\worldf$\\
 $\bm{\omega}_{\id}\in\realnum{3}$, $l_{\id}\in\realnum{}$&angular velocity, length of $\id^{th}$ cable\\
 $\vecrho{\id}\in\realnum{3}$&position of $\id^{th}$ attach point in $\loadf$\\
    \hline\hline
\end{tabularx}
%\begin{tabbing}
%$^{\mathrm{a}}$Ref.~\cite{varshni90}. \hspace{25pt} \= 
%$^{\mathrm{b}}$Ref.~\cite{demiralp05}. \hspace{25pt} \=
%$^\dag$PR implies Present Result.
%\end{tabbing}
\end{table}

\section{System Modeling}\label{sec:modeling}

\begin{figure}[t] 
\centering 
\includegraphics[width=0.9\columnwidth, trim=130 85 80 50, clip]{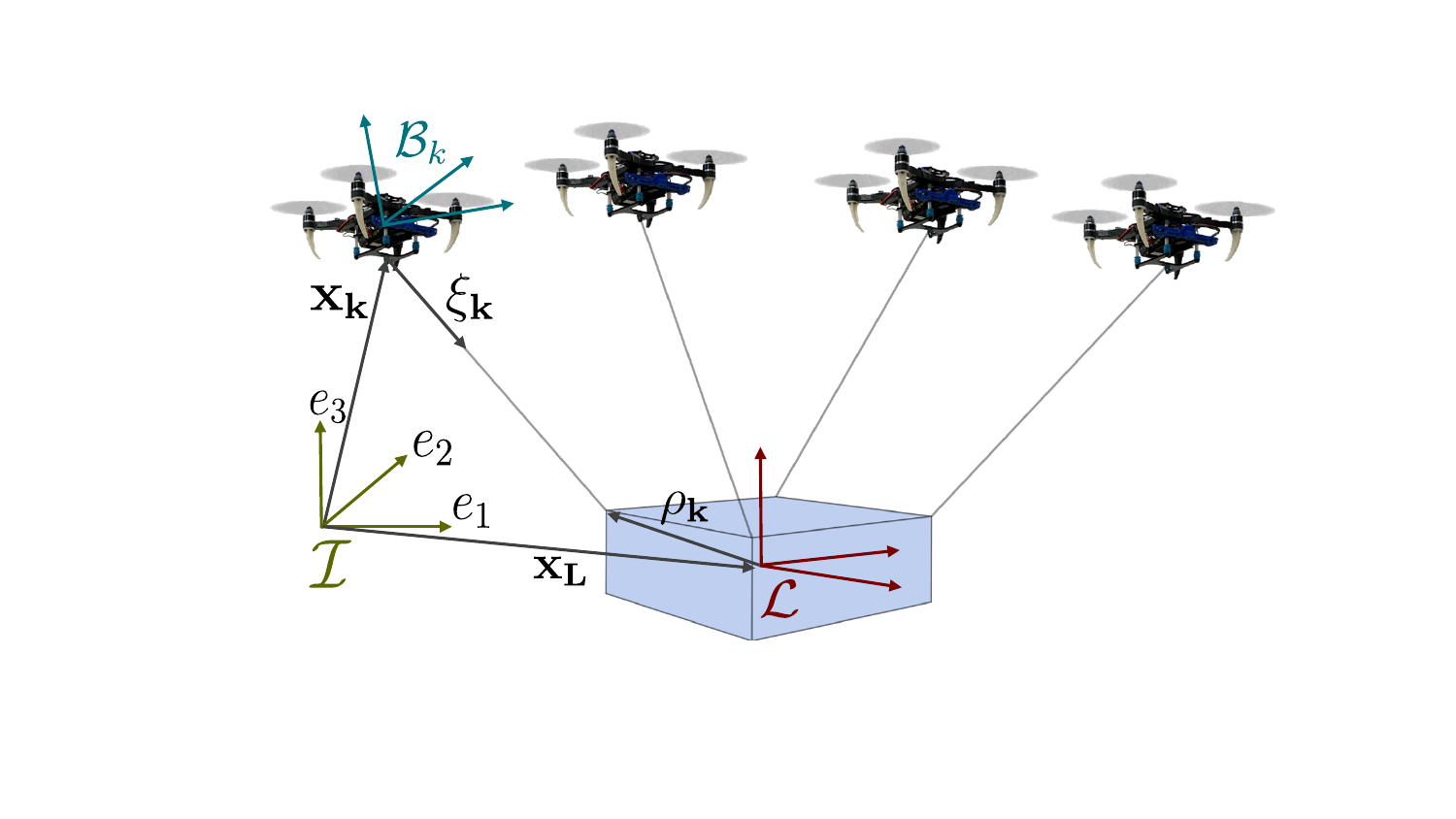}
\caption{System convention definition: $\worldf$, $\loadf$, $\robotf{\id}$ denote the world frame, the payload body frame, and the $\id^{th}$ robot body frames, respectively, for a generic quadrotor team that is cooperatively transporting and manipulating a cable-suspended payload.} \label{fig:frame_definition} 
\vspace{-20pt}
\end{figure}

In this section, we formulate the system dynamics of $n$ quadrotor manipulating a rigid-body payload via suspended cables, as illustrated in Fig.~\ref{fig:frame_definition}. The model introduced in this section will serve as the foundation for our trajectory planning method, which aims to generate collision-free trajectories for both the payload, the cables, and the quadrotors as they navigate through cluttered environments.

We begin this section by outlining the assumptions underlying our dynamics model. Subsequently, we derive the system dynamics utilized in the trajectory planning algorithm. The notation used in this section and the rest of the paper is summarized in Table~\ref{tab:notation}.

The dynamics model is based on the following assumptions:

\begin{enumerate} 
\item Aerodynamic drag on the payload and quadrotors are considered negligible due to the low-speed maneuvers involved, simplifying the dynamic equations. 
\item The cables are assumed to be massless and are attached directly to the quadrotors' centers of mass, eliminating the need to account for cable mass distribution. 
\item Aerodynamic effects among the robots and the payload are neglected because their influence is minimal in the operational scenarios considered.
\item All the cables are taut during the transportation and manipulation tasks.
\end{enumerate}

% \begin{figure}[htbp] \centering \includegraphics[width=0.5\textwidth]{figs/multi-robot-cable-frame.png} \caption{System convention definition: $\worldf$, $\loadf$, $\robotf{\id}$ denote the world frame, the payload body frame, and the $\id^{th}$ robot body frames, respectively, for a generic quadrotor team that is cooperatively transporting and manipulating a cable-suspended payload.} \label{fig:frame_definition} 
% \end{figure}

We first define the payload pose vector, payload velocity vector, and its corresponding derivative as 
\begin{equation}
    \matX_L = \begin{bmatrix}
        \loadpos\\\loadquat
    \end{bmatrix},\,\matV_L=\begin{bmatrix}
        \loadvel\\\loadangvel
    \end{bmatrix},\,\dot{\matV}_L=\begin{bmatrix}
        \loadacc\\\loadangacc
    \end{bmatrix},
\end{equation}
where $\loadpos,\loadvel,\loadacc$ are the payload's position, velocity, and acceleration with respect to $\worldf$,  $\loadquat$ is the quaternion representation of the payload's orientation with respect to $\worldf$, and $\loadangvel,\loadangacc$ are the payload's angular velocity and acceleration with respect to the payload frame $\loadf$.

By differentiating both sides of the equation of $\matX_L$, we can obtain the kinematics model of the payload as 
\begin{equation}
\dot{\matX}_L = \begin{bmatrix}
        \loadvel\\\loadquatdot
    \end{bmatrix} = \begin{bmatrix}
        \loadvel\\\half\loadquat\otimes\loadangvel
    \end{bmatrix},
    \label{eq:payload_kinematics}
\end{equation}
where $\otimes$ denotes quaternion multiplication.
Further, since we model the payload as a rigid body, we can easily obtain the payload's equations of motion as 
\begin{equation}
\begin{split}
\loadmass\loadacc{} = \netplforce-\loadmass\vecg,\hspace{0.8em} 
\inertiaload\loadangacc &= \netplmoment-\loadangvel\times\inertiaload\loadangvel,
\label{eq:payload_dynamics}
\end{split}
\end{equation}
where $\netplforce$ is the total force acting on the payload by the cables in $\worldf$ and $\netplmoment$ is the total moments generated by the cable forces on the payload in $\loadf$. $\netplforce$ and $\netplmoment$ can be expressed as functions of the cable forces
\begin{equation}
\netplforce=\sum_{\id=1}^{n}\tension{\id},\,~\netplmoment=\sum_{\id=1}^{n}\vecrho{\id}\times\loadrot^{\top}\tension{\id},
\label{eq:payload_wrench}
\end{equation}
where $\tension{\id}$ denotes the $\id^{th}$ cable tension force, expressed in the inertia frame $\worldf$, $\loadrot$ is the rotation matrix that represents the payload's orientation with respect to the inertia frame $\worldf$, and $\vecrho{\id}$ is the position vector of the $\id^{th}$ attach point corresponds to the $\id^{th}$ cable and quadrotor in the payload frame $\loadf$. 

For the convenience of the derivation of the later part of this section for the planning, we rotate the force $\matF$ into the frame $\loadf$ as 
\begin{equation}
    \matF^{\loadf} = \loadrot^\top\matF = \sum_{\id=1}^{n}\tension{\id}^{\loadf},\,\,\, \tension{\id}^{\loadf} = \loadrot^\top\tension{\id}, \label{eq:payload_force_loadf}
\end{equation}
where $\tension{\id}^{\loadf}$ denotes the $\id^{th}$ cable tension force, expressed in the load frame $\loadf$.

By writing eq.~(\ref{eq:payload_wrench}) and (\ref{eq:payload_force_loadf}) in matrix form, we obtain the wrench $\matW^{\loadf}$ in the payload frame $\loadf$ as:

\begin{equation} \matW^{\loadf} = \begin{bmatrix} \matF^{\loadf} \\ \matM \end{bmatrix} = \matP \tension{}^{\loadf},\,\,\, \tension{}^{\loadf} = \begin{bmatrix} \tension{1}^{\loadf} \\ \vdots \\ \tension{n}^{\loadf}, \end{bmatrix} \label{eq:wrench_matrix_form} \end{equation}
where $\matP \in \mathbb{R}^{6 \times 3n}$ maps the tension forces of all $n$ quadrotors to the wrench on the payload. The mapping matrix $\matP$ is constructed as:

\begin{equation} \matP = \begin{bmatrix} \matI_{3} & \cdots & \matI_{3} \\ \hatvecrho{1} & \cdots & \hatvecrho{n} 
\end{bmatrix} \label{eq:P_matrix}, 
\end{equation}
where $\matI_{3}$ is the $3 \times 3$ identity matrix, and $\hatvecrho{k}$ denotes the skew-symmetric matrix of $\vecrho{k}$, defined such that $\hatvecrho{k} \vecb = \vecrho{k} \times \vecb$ for any $\vecb \in \mathbb{R}^3$. With this model, if a planner computes the forces $\matF$ and moments $\matM$ acting on the payload, we can further determine the tension forces in each cable using the following equation:
\begin{equation}
    \tension{}^{\loadf} = \matP^\top(\matP\matP^\top)^{-1}\matW^{\loadf}.
    \label{eq:inverse_tension_distribution}
\end{equation}

On the other hand, assuming the cables remain taut as shown in Fig.~\ref{fig:frame_definition}, we can express the position of the $k^{th}$ quadrotor relative to the payload frame $\loadf$ as:
\begin{equation} 
    \robotpos{k}^{\loadf} = \vecrho{k} - l_k\cablevec{k}^\loadf,\label{eq:robot_pos_loadf} 
\end{equation}
where $\cablevec{k}^\loadf \in \mathbb{R}^3$ is the unit vector representing the direction of the $k^{th}$ cable the $k^{th}$ quadrotor to the corresponding attachment point on the payload, and $l_k$ denotes the length of the $k^{th}$ cable. Since the cable force $\tension{k}^{\loadf}$ is along the cable, we can further derive the unit vector as:
\begin{equation}
    \cablevec{k}^\loadf = -\frac{\tension{k}^{\loadf}}{\twonorm{\tension{k}^{\loadf}}},\,\,\,\cablevec{\id} = \loadrot\cablevec{\id}^\loadf.\label{eq:cable_direction_uniform_cable_force}
\end{equation}
Using eq.~(\ref{eq:robot_pos_loadf}), the position of the $k^{th}$ quadrotor in the inertial frame $\worldf$ can be expressed as:
\begin{equation} 
    \robotpos{k} = \loadpos + \loadrot \robotpos{k}^{\loadf},\label{eq:robot_pos_worldf} 
\end{equation}

As we observe the above equations, a key characteristic of this system is that the direction of each cable dictates the corresponding cable forces acting on the payload. These forces, in turn, influence both the total force $\matF$ and moments $\matM$ on the payload. As a result, planning the motion of the payload, and consequently the forces $\matF$ and moments $\matM$, indirectly determines the cable directions and thus the positions of the quadrotors. This interdependence is critical, as the quadrotors and the cables must avoid obstacles while ensuring safe and efficient manipulation of the payload. This relationship affects how we define the state space we plan within, which is integral to the trajectory planning algorithm introduced in Section~\ref{sec:trajectory_generation}.

%% file: sections/04-Planning.tex
\begin{figure}[t!]
\centering
  \includegraphics[width=0.9\columnwidth]{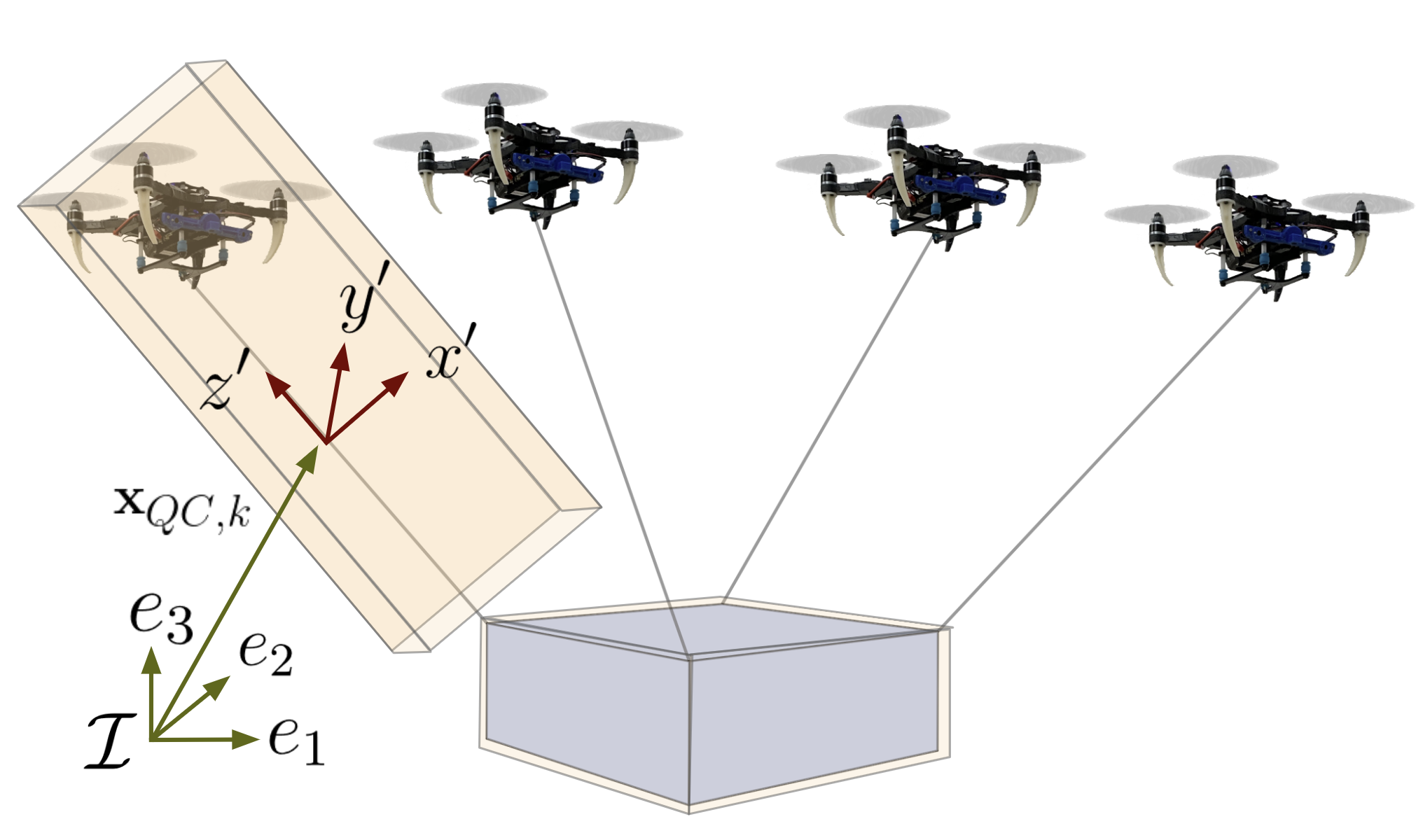}  
  \caption{Polytopic approximation of the payload, cables, and quadrotors.\label{fig:polytope}}
  \vspace{-20pt}
\end{figure}

\section{Planning}\label{sec:trajectory_generation}

In this section, we present an optimal trajectory planning method for a team of quadrotors that cooperatively manipulate a cable-suspended payload. The planning method accounts for the complex system dynamics and ensures that the entire system, including the quadrotors, cables, and payload, avoids obstacles in the environment.

First, we model the payload, quadrotors, suspended cables, and obstacles in the environment as convex 3D polytopes, as depicted in Fig.~\ref{fig:polytope}. Each quadrotor and its corresponding cable are encapsulated within a single polytope. We then formulate the trajectory planning problem as an optimization task, which navigates the quadrotor-payload system through a cluttered environment from an initial pose to a desired goal pose. Obstacle avoidance is achieved through the use of Control Barrier Functions (CBFs) integrated as constraints within the optimization problem. Additionally, the proposed method respects the system's nonlinear dynamics, ensuring feasible and collision-free trajectories.

\subsection{System States and Inputs}

We define the state vector $\matX$ and input vector $\matU$ for the system as follows:
\begin{equation*} 
    \matX = \begin{bmatrix}
        \loadpos^\top, \loadquat^\top, \loadvel^\top, \loadangvel^\top, \matF^\top, \matM^\top
    \end{bmatrix}^\top,
   \matU = \begin{bmatrix}
        \dot{\matF}^{\top}, \dot{\matM}^{\top}
    \end{bmatrix}^\top, 
\end{equation*}
where $\loadpos$ and $\loadquat$ represent the payload's position and orientation, $\loadvel$ and $\loadangvel$ are the linear and angular velocities of the payload, and $\matF$ and $\matM$ are the total forces and moments acting on the payload. The inputs $\dot{\matF}$ and $\dot{\matM}$ are the time derivatives of the forces and moments.

\subsection{Optimization Problem Formulation}

Our planning algorithm optimizes over a sequence of system states $\{\matX_0,\matX_1,\cdots,\matX_N\}$ and inputs $\{\matU_0,\matU_1,\cdots,\matU_{N-1}\}$ over a fixed time horizon. The objective function includes both a running cost $h(\matX,\matU)$ and a terminal cost $h_N(\matX)$, and is formulated as follows:
\begin{subequations}
    \label{eq:MPC_Form}
    \begin{align}
        \min_{\substack{\matX_0,\cdots,\matX_N, \\ \matU_0,\cdots,\matU_{N-1}}}& ~\sum_{i=0}^{N-1}  h\left(\matX_i,\matU_i\right) + h_N\prths{\matX_N}\\ 
        \text{subject to:}&~\matX_{i+1} = \mathbf{f}\left(\matX_i,\matU_i\right),\\
        &~\matX_0 = \matX(t_0), ~\mathbf{g}\left(\matX_i,\matU_i\right)\leq 0,
    \end{align}
\end{subequations}
where $\matX_{i+1} = \mathbf{f}\prths{\matX_i,\matU_i}$ represents the system's nonlinear dynamics in discrete form, and $\mathbf{g}\left(\matX_i,\matU_i\right)$ includes additional state and input constraints. The optimization process begins from the initial condition $\matX_0$, while ensuring the system dynamics $\mathbf{f}(\matX, \matU)$ are respected.

\subsection{Cost Function}

The objective of this task is to manipulate the payload to reach a target pose. To achieve this, we define the following objective function:
\begin{equation}
    \vece_{\matX_N}^\top\matQ_{X_N}\vece_{\matX_N} + \sum_{i=0}^{N-1} \left( \vece_{\matX_i}^\top\matQ_X\vece_{\matX_i} + \vece_{\matU_i}^\top\matQ_U\vece_{\matU_i} \right),
\label{eq:nmpc_costs}
\end{equation}
where $\vece_{\matX_i}$ and $\vece_{\matU_i}$ represent the state and input errors between the predicted and desired values at each time step. The goal pose of the payload can serve as the reference for all states along the trajectory horizon. However, this may cause the planner to get stuck in local minima. To mitigate this, we can also employ a simple global planner, such as the $A^{*}$ algorithm, to compute a reference path for the payload's position. Our proposed trajectory planner then refines this path by generating a complete trajectory for all payload states, including both poses and twists.

\subsection{CBF Constraint Formulation for Obstacle Avoidance}

Obstacle avoidance is incorporated into the control problem using Control Barrier Functions (CBFs). Unlike traditional distance-based constraints, which are only triggered when obstacles intersect the planning horizon, CBFs provide a more computationally efficient solution, preventing collisions proactively.

We use discrete-time CBF constraints in the following form~\cite{ames_control_2019}:
\begin{equation}
    h(\matX_{i+1}) \geq \gamma_i h(\matX_i),\,\,\, 0\leq\gamma_i <1\label{eq:CBF}
\end{equation}
where $h(\matX)\geq 0$ is a CBF measuring the distance between the system (payload, cables, and quadrotors) and the obstacles. For example, the CBF for the payload can be formulated as:
\begin{subequations}
\begin{align} 
    h(\matX) &= \min_{\vecy_L,\vecy_{obs}} \twonorm{\vecy_L - \vecy_{obs}}^2 - d_{safe}^2, \\
   \text{s.t.} \quad & \matA_{L}(\matX) \vecy_{L} \leq \matB_{L}(\matX),\,\,\, \matA_{obs} \vecy_{obs} \leq \matB_{obs},
\end{align}
   \label{eq:min_dist_bw_obs_load}
\end{subequations}
where $d_{safe}$ is the safe distance between the payload polytope and obstacle polytopes, $\vecy_L$ and $\vecy_{obs}$ of any points in the payload and obstacle polytopes, respectively, and $\matA_L$ and $\matA_{obs}$ represent the geometric parameters of the polytopes. For quadrotor-cable pairs, the same formulation applies, but the subscripts $*_L$ are substituted with $*_{QC}$.

\subsubsection{Dual Formulation for CBF Constraints}

To avoid solving the optimization problem in eq.~(\ref{eq:min_dist_bw_obs_load}) directly, we apply the Duality theorem to derive equivalent CBF constraints. For the payload CBF, this is formulated as~\cite{boyd2004convex}:
\begin{subequations}
\begin{align}
 g(\matX)&=\max_{\lambda^{L},\lambda^{obs}}-\lambda^{obs}\matB_{obs}-\lambda^L\matB_L(\matX), \\
\text{s.t.} \quad &\lambda^{obs}\matA_{obs}+\lambda^{L}\matA_{L}(\matX)=0, \\
                  &\lambda^{obs} \geq 0,~ \lambda^{L} \geq 0,~  \twonorm{\lambda^{obs}*A_{obs}} \leq 1.
\end{align}
\end{subequations}
Since the original problem is convex with linear constraints, strong duality holds, leading to $g(\matX) = h(\matX)$. Additionally, applying the weak duality theorem gives:
\begin{equation}
    -\lambda^{obs}\matB_{obs}-\lambda^L\matB_L(\matX) \leq g(\matX) = h(\matX). \label{eq:weak_duality}
\end{equation}

From eq.~(\ref{eq:weak_duality}), we derive the stronger CBF constraints:
\begin{subequations}
\begin{align}
    &-\lambda_i^{obs}\matB_{obs}-\lambda_i^L\matB_L(\matX_{i+1}) \geq \gamma_ih(\matX_i), \label{eq:stronger_cbf}\\
    &\lambda_i^{obs}\matA_{obs}+\lambda_i^{L}\matA_{L}(\matX_{i+1})=0, \\
    &\lambda_i^{obs} \geq 0,~\lambda_i^{L} \geq 0,~  \twonorm{\lambda_i^{obs}*A_{obs}} \leq 1.
\end{align}
\end{subequations}
\subsubsection{Exponential CBF Constraints}

To further reduce computational complexity, we apply exponential CBF constraints~\cite{ames_control_2019}, along with a relaxation variable $\alpha_i$~\cite{zeng2021cdc}, modifying eq.~(\ref{eq:stronger_cbf}) to:
\begin{equation}
    -\lambda_i^{obs}\matB_{obs}-\lambda_i^L\matB_L(\matX_{i+1}) \geq \alpha_i\Pi_{j=0}^i\gamma_jh(\matX_0). \label{eq:exponential_stronger_cbf}
\end{equation}
Here, $\alpha_i$ is optimized alongside other variables in eq.~(\ref{eq:MPC_Form}), and $h(\matX_0)$ is computed at each initial condition.

\subsubsection{Representing Objects as Convex Polytopes}

We now elaborate on constructing convex polytopes for system components (payload, quadrotor-cable pairs) and the associated CBF constraints.

For each polytope, we define a local body frame and represent the polytope as $\mathbf{A}_b\vecy_b\leq\mathbf{B}_b$, where $\matA_b$ and $\matB_b$ are constant matrices and vectors derived from the object's geometry. The payload polytope is defined in the local payload frame $\loadf$, centered at the payload’s center of mass.

For each quadrotor and cable pair, we define a local frame $\mathcal{QC}_\id$, whose origin is at the center of the $\id^{th}$ cable (see Fig.~\ref{fig:polytope}). The $z$-axis aligns with the cable, the $y'$-axis is defined as $(z' \times \axis{1}{})$, and the $x'$-axis is obtained as $y' \times z'$. As we defined the frame $\mathcal{QC}_\id$, the rotation matrix from $\mathcal{QC}_\id$ to $\worldf$ and the position of $\mathcal{QC}_\id$'s origin with respect to $\worldf$ are 
\begin{equation}
\begin{split}
 \mathbf{R}_{QC,k} &= \begin{bmatrix}
     (\axis{1}{}\times\cablevec{\id}) \times \cablevec{\id} &     (-\axis{1}{}\times\cablevec{\id}) & -\cablevec{\id} 
 \end{bmatrix},\\
  \vecx_{QC,\id} &= \loadpos + \loadrot \vecrho{\id}- l_\id\cablevec{\id}/2.
\end{split}
\end{equation}

The quadrotor-cable pair is approximated by a cuboid polytope, with dimensions based on the length of the cable and the size of the quadrotor.
Given a map of the environment, with convex obstacles expressed in the inertial frame as $\matA_{obs}\vecy_{obs}\leq\matB_{obs}$, we transform the polytopes into a common reference frame for constraint computation. For any obstacle point $\vecy_{obs}$, we have:
\begin{equation}
    \vecy_{obs} = \loadrot\vecy_{obs}^{\loadf} + \loadpos 
    = \matR_{QC,\id}\vecy_{obs}^{\mathcal{QC}_\id} + \vecx_{QC,\id}.\label{eq:transformed_polytope_pts}
\end{equation}
Substituting eq.~(\ref{eq:transformed_polytope_pts}) into $\matA_{obs}\vecy_{obs}\leq\matB_{obs}$, we derive:
\begin{equation}
    \matA_{obs}^{\loadf}\vecy_{obs}^{\loadf} \leq\matB_{obs}^{\loadf}, \,\,\,\matA_{obs}^{\mathcal{QC}_\id}\vecy_{obs}^{\mathcal{QC}_\id} \leq\matB_{obs}^{\mathcal{QC}_\id},
\end{equation}
where $\matA_{obs}^{\loadf} = \matA_{obs}\loadrot, \matA_{obs}^{\mathcal{QC}_\id} = \matA_{obs}\loadrot, \matB_{obs}^{\loadf}=\matB_{obs} - \matA_{obs}\vecx_{QC,\id}, \matB_{obs}^{\mathcal{QC}_\id}=\matB_{obs} - \matA_{obs}\loadpos$.
We then substitute $\matB_{obs}$ in eq.~(\ref{eq:stronger_cbf}) with $\matB_{obs}^{\loadf}$ and $\matB_{obs}^{\mathcal{QC}_\id}$ to fully define the inequality constraints.

%% file: sections/05-Experiments.tex
%!TEX root = ARTICLE.tex
\section{Experimental Results}~\label{sec:experimental_results}
\vspace{-10pt}
% Include Figure here
\begin{figure}[t]
    \centering
    % Replace 'figure_image.png' with your actual figure file name and add appropriate caption and label
    \includegraphics[width=\columnwidth]{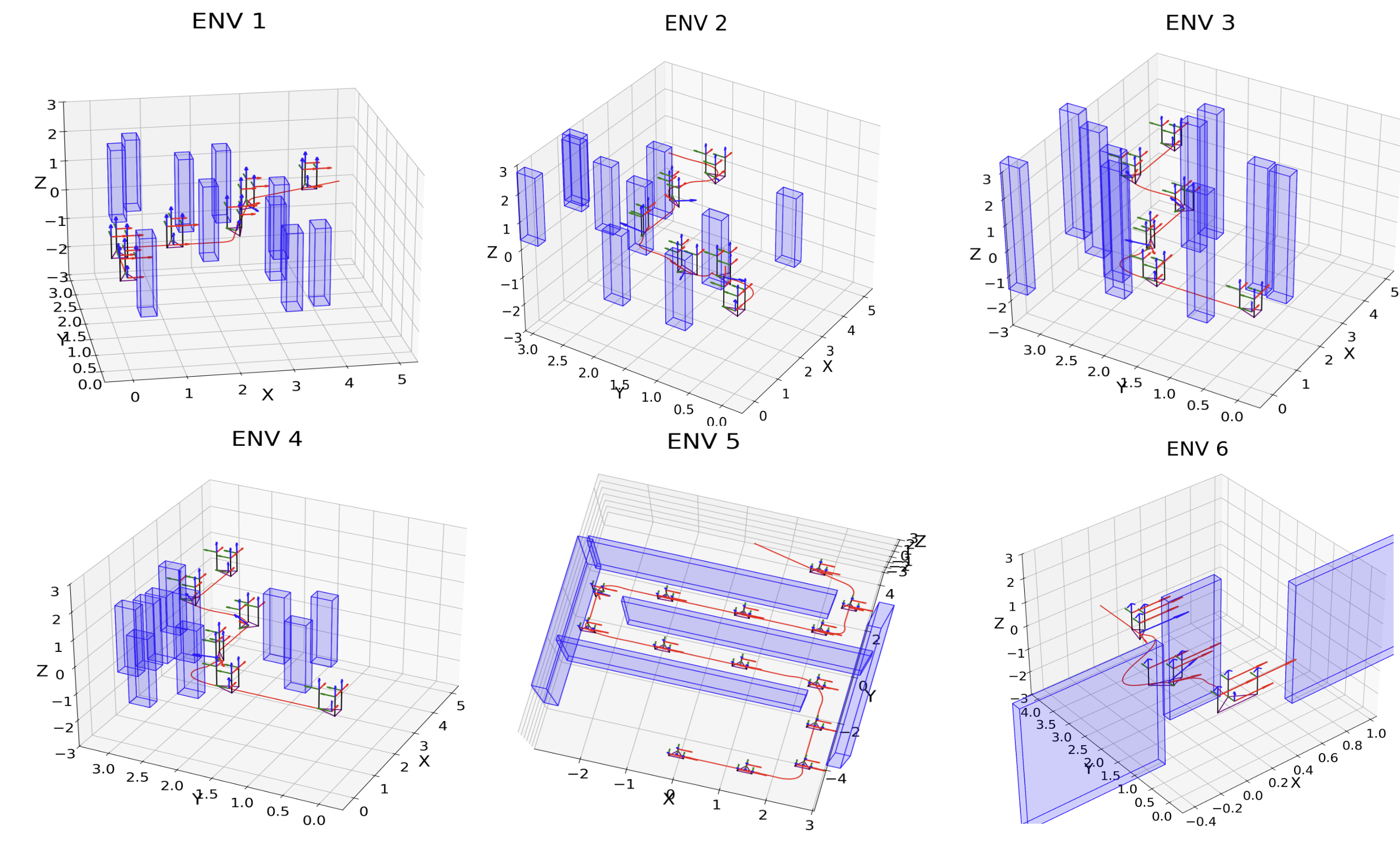}
    \caption{Sample trajectory generation in sim environments. Every 80th point is shown with axis for clarity .\label{fig:sim_envs}}
    \vspace{-10pt}
\end{figure}
\begin{figure}[t]
    \centering
    \includegraphics[width=\columnwidth
]{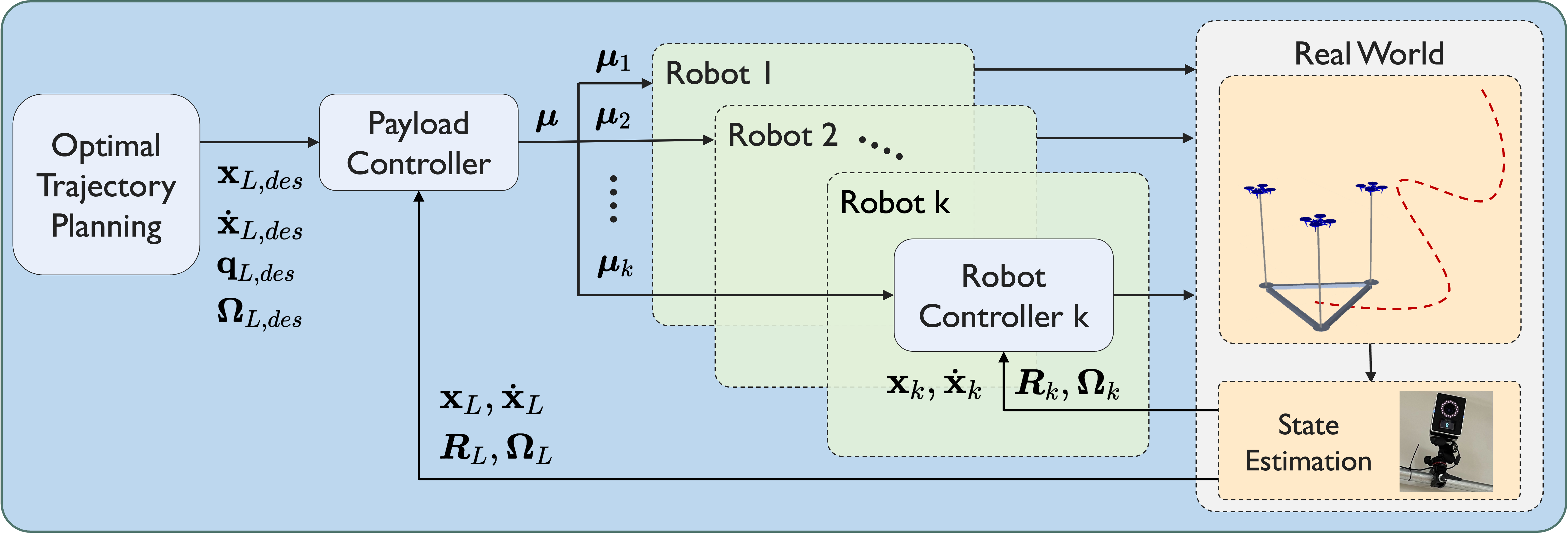}
    \caption{System block diagram in the real-world experiments.}
    \vspace{-10pt}
    \label{fig:block-diagram}
\end{figure}

In this section, we discuss the experiments we conducted in simulator and the real world to validate our proposed trajectory planner. 

\subsection{Setup}

For Simulation, we employ RotorTM~\cite{guanrui2024rotortm}, an open source simulator for multi-robot transportation. This allows us to determine the limits on state and control input for the trajectory planner. We also design more complex environment in the simulator to test our trajectory planner.

We conducted real-world experiments in a $10\times6\times4~\si{m}^3$ flying space at the ARPL lab at New York University, using a Vicon motion capture system for the feedback of the payload, attach points, and quadrotors at $100~\si{Hz}$. The geometric controller from our previous work~\cite{guanrui2021iser} controls the system, using the desired payload state from the proposed planning algorithms. The controller distributes cable tension forces to each quadrotor, and the onboard controllers generate motor commands to achieve the required tensions. Fig.~\ref{fig:block-diagram} illustrates the data flow between the motion capture system, planning algorithm, controller, and quadrotors.

In the real-world experiments, we create a narrow gap using termocol boxes. We use the planner to find a path through the gap. We increase the size of the obstacle by $7.5~\si{cm}$ before feeding into the planner to account for uncertainty in the real-world system.

\begin{table}[t]
\centering
\begin{tabular}{|c|c|c|}
\hline
\textbf{Environments} & \textbf{Without Global Planner} & \textbf{With Global Planner} \\ \hline
ENV 1    & 60\%                           & 100\%                       \\ \hline
ENV 2    & 60\%                           & 100\%                       \\ \hline
ENV 3    &   50\%                         & 100\%                   \\ \hline
ENV 4    &   70\%                         & 100\%                   \\ \hline
ENV 5    &   0\%                          & 100\%                   \\ \hline
ENV 6    &   20\%                         & 100\%                   \\ \hline
\end{tabular}
\caption{Result in simulation environment. Start points and goal points were chosen randomly in a given range to check the robustness of the planner. Obstacles in environment 1-4 were randomly generated.\label{tab:results_success_rates}}
\vspace{-20pt}
\end{table}

\begin{figure*}
    \centering
    % Replace 'figure_image.png' with your actual figure file name and add appropriate caption and label
    \includegraphics[width=0.9\textwidth]{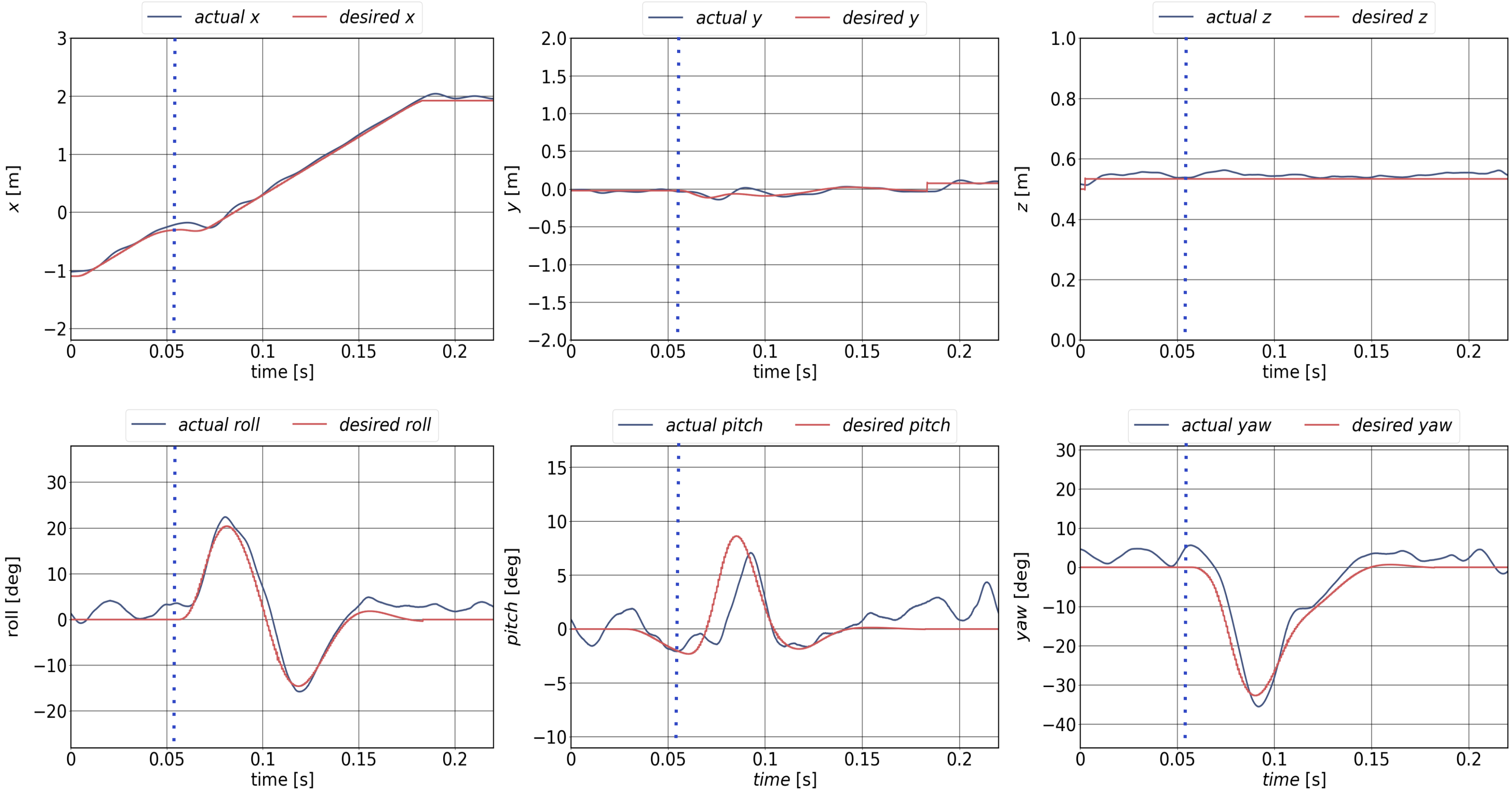}
    \caption{Trajectory tracking of a cable-suspended payload manipulated by three quadrotors in a real-world environment. The plot showcases the path of the payload as it navigates adjusting its orientation and position to avoid obstacles. Dotted Line corresponds to the time at which the payload starts entering the tight gap.\label{fig:null_space_visualization}}
\end{figure*}
\subsection{Trajectory Generation}

We tested our algorithm on a triangular payload, weighted $196~\si{g}$, carried by $3$ robots attached to its $3$ corners via cables. We approximate our payload, cables, and quadrotors as convex polytopes. In our experiments, we had an equilateral triangular payload on side $1 \si{m}$, so we took a polytope with the top and bottom faces of the same dimensions as the triangle dimensions and a height of $0.1~\si{m}$. A plane in the form of $\matA\mathbf{y} \leq \matB$, where $\mathbf{y} \in \realnum{3}$ divides the 3D space into two parts. So, our polytope which has five planes can be combined together to represent all points in and on the polytope as $\matA\mathbf{y} \leq \matB$, where $\mathbf{y} \in \realnum{3}$. The cables and cuboid are similarly approximated together as cuboids of dimensions $0.15~\si{m} \times 0.15~\si{m} \times 1~\si{m}$. Similarly, we also model the obstacles in the environment as convex polytopes.
 We formulate the optimization problem in the Casadi optimizer with an horizon length of $25$ steps. We take the first input, apply on the system and then re-solve the optimization problem. Each iteration takes between $9$ and $15~\si{s}$ on an Intel(R) Core(TM) i7-8565U CPU. The time depends on the number of obstacles that are active in a given state. We chose to ignore obstacles farther than $0.6~\si{m}$ in our planning. 

\subsection{Results}
First, we report the simulation results. In the simulation, we conducted in complex environments with multiple obstacles presented, as shown in Fig.~\ref{fig:sim_envs}. We can observe that the quadrotors performed aggressive braking to prevent the payload from colliding with walls, while simultaneously moving as quickly as possible within the given constraints. The quadrotors also collaborated to rotate the payload around each axis while navigating through the environments.

To assess the robustness of our planner, we also tested it in a variety of simulated environments with 10 different randomly generated pairs of start and goal poses within each environment. We report the success rates of the planning algorithms, which are summarized in Table~\ref{tab:results_success_rates}.

In larger and twisted environments, we observed that our proposed planners often stuck in locally optimal solutions or reach states from which recovery was infeasible if we don't use global planner to provide some sorts of references. When integrating a simple global planner, we mitigated these issues by providing a rough reference path for the system to follow, allowing our proposed algorithms to plan more effectively with a success rate of $100\%$.

In real-world experiments, we observed that the quadrotors manipulate the payload by rotating significantly to navigate through the gap between two columns. This maneuver allowed the system to pass through the obstacles. The quadrotors demonstrated coordinated behavior, squeezing through obstacle columns and spreading out after clearing them. These results highlight that the generated trajectory successfully ensures that both the payload and quadrotors avoid collisions throughout the task.

%% file: sections/06-Conclusion.tex
%!TEX root = ARTICLE.tex
\section{Conclusion}~\label{sec:conclusion}
This paper presents a comprehensive trajectory planning algorithm for multi-quadrotor systems manipulating a cable-suspended payload in cluttered environments. By incorporating control barrier functions (CBFs) into the planning process, the proposed method ensures obstacle avoidance for the payload, cables, and quadrotors, effectively addressing the challenges posed by the complex dynamics of the system. We use convex polytopes to model the system components and apply the Duality theorem to reduce the computational complexity of the optimization process, making it feasible for real-world applications. Experimental results in both simulation and real-world environments validate the proposed method's ability to generate collision-free trajectories.

Future works will explore further optimization techniques and extend the framework to accommodate more dynamic and unpredictable environments, improving the system's robustness and applicability to a wider range of aerial manipulation tasks.